\documentclass[10pt,twocolumn,letterpaper]{article}

\usepackage[pagenumbers]{iccv} 

\usepackage{arydshln}
\usepackage{colortbl}
\usepackage{diagbox}
\usepackage{gensymb}
\usepackage{graphicx}
\usepackage{mathtools}
\usepackage{microtype}
\usepackage{multirow}
\usepackage{nicefrac}
\usepackage{scalerel}
\usepackage{siunitx}
\usepackage{soul}
\usepackage{textcomp}
\usepackage[normalem]{ulem}

\colorlet{first}{Green!30}       
\colorlet{second}{LimeGreen!30}   
\colorlet{third}{Yellow!30}      
\newcommand{\first}{\cellcolor{first}\bf}   
\newcommand{\second}{\cellcolor{second}}      

\definecolor{darkgreen}{rgb}{0.0,0.6,0.4}
\newcommand{\rd}{\textcolor{red}}
\newcommand{\gr}{\textcolor{darkgreen}}

\newcommand\blfootnote[1]{
    \begingroup
    \renewcommand\thefootnote{}\footnote{#1}
    \addtocounter{footnote}{-1}
    \endgroup
}

\definecolor{iccvblue}{rgb}{0.21,0.49,0.74}
\usepackage[pagebackref,breaklinks,colorlinks,allcolors=iccvblue]{hyperref}

\def\paperID{6901}
\def\confName{ICCV}
\def\confYear{2025}

\title{
    CasP: Improving Semi-Dense Feature Matching Pipeline Leveraging\\
    Cascaded Correspondence Priors for Guidance
}

\author{
    {Peiqi Chen}$^{1\star}$ \hspace*{0.5em}
    {Lei Yu}$^{2\star}$ \hspace*{0.5em}
    {Yi Wan}$^{1\dagger}$ \hspace*{0.5em}
    {Yingying Pei}$^{1}$ \hspace*{0.5em}
    {Xinyi Liu}$^{1}$ \hspace*{0.5em}
    {Yongxiang Yao}$^{1}$ \\
    {Yingying Zhang}$^{2}$ \hspace*{0.5em}
    {Lixiang Ru}$^{2}$ \hspace*{0.5em}
    {Liheng Zhong}$^{2}$ \hspace*{0.5em}
    {Jingdong Chen}$^{2}$ \hspace*{0.5em}
    {Ming Yang}$^{2}$ \hspace*{0.5em}
    {Yongjun Zhang}$^{1\dagger}$ \\
    $^{1}${Wuhan University} \quad\quad
    $^{2}${Ant Group} \\
}

\begin{document}
    \maketitle
    \blfootnote{
        $^{\star}$Equal contribution.
        $^{\dagger}$Corresponding author.
    }

    \begin{abstract}
Semi-dense feature matching methods have shown strong performance in challenging scenarios. However, the existing pipeline relies on a global search across the entire feature map to establish coarse matches, limiting further improvements in accuracy and efficiency. Motivated by this limitation, we propose a novel pipeline, CasP, which leverages cascaded correspondence priors for guidance. Specifically, the matching stage is decomposed into two progressive phases, bridged by a region-based selective cross-attention mechanism designed to enhance feature discriminability. In the second phase, one-to-one matches are determined by restricting the search range to the one-to-many prior areas identified in the first phase. Additionally, this pipeline benefits from incorporating high-level features, which helps reduce the computational costs of low-level feature extraction. The acceleration gains of CasP increase with higher resolution, and our lite model achieves a speedup of $\sim2.2\times$ at a resolution of 1152 compared to the most efficient method, ELoFTR. Furthermore, extensive experiments demonstrate its superiority in geometric estimation, particularly with impressive cross-domain generalization. These advantages highlight its potential for latency-sensitive and high-robustness applications, such as SLAM and UAV systems. Code is available at \url{https://github.com/pq-chen/CasP}.
\end{abstract}
    
    \section{Introduction}
\label{sec:introduction}

Local feature matching is a fundamental task in 3D computer vision that aims to establish correspondences within each image pair. This technique is crucial for accurate geometric estimation and supports a wide range of downstream applications, including structure-from-motion~\cite{schonberger2016structurefrommotion,lindenberger2021pixelperfect,he2024detectorfree} and visual localization~\cite{sarlin2019coarse,toft2020longterm,sarlin2021back}. In particular, real-time processing tasks, such as SLAM and UAV systems, demand high computational efficiency and robustness. The classical pipeline adopted by sparse methods consists of feature detection followed by feature description. However, its success largely depends on the detector’s ability~\cite{lowe2004distinctive,tyszkiewicz2020disk,detone2018superpoint} to generate repeatable key points. Despite proposals to employ graph neural networks~\cite{sarlin2020superglue,lindenberger2023lightglue} to enhance off-the-shelf local features, reliable detection remains unguaranteed, especially in areas with low texture and repetitive patterns.

\begin{figure}[!t]
\centering
\includegraphics[width=0.95\linewidth]{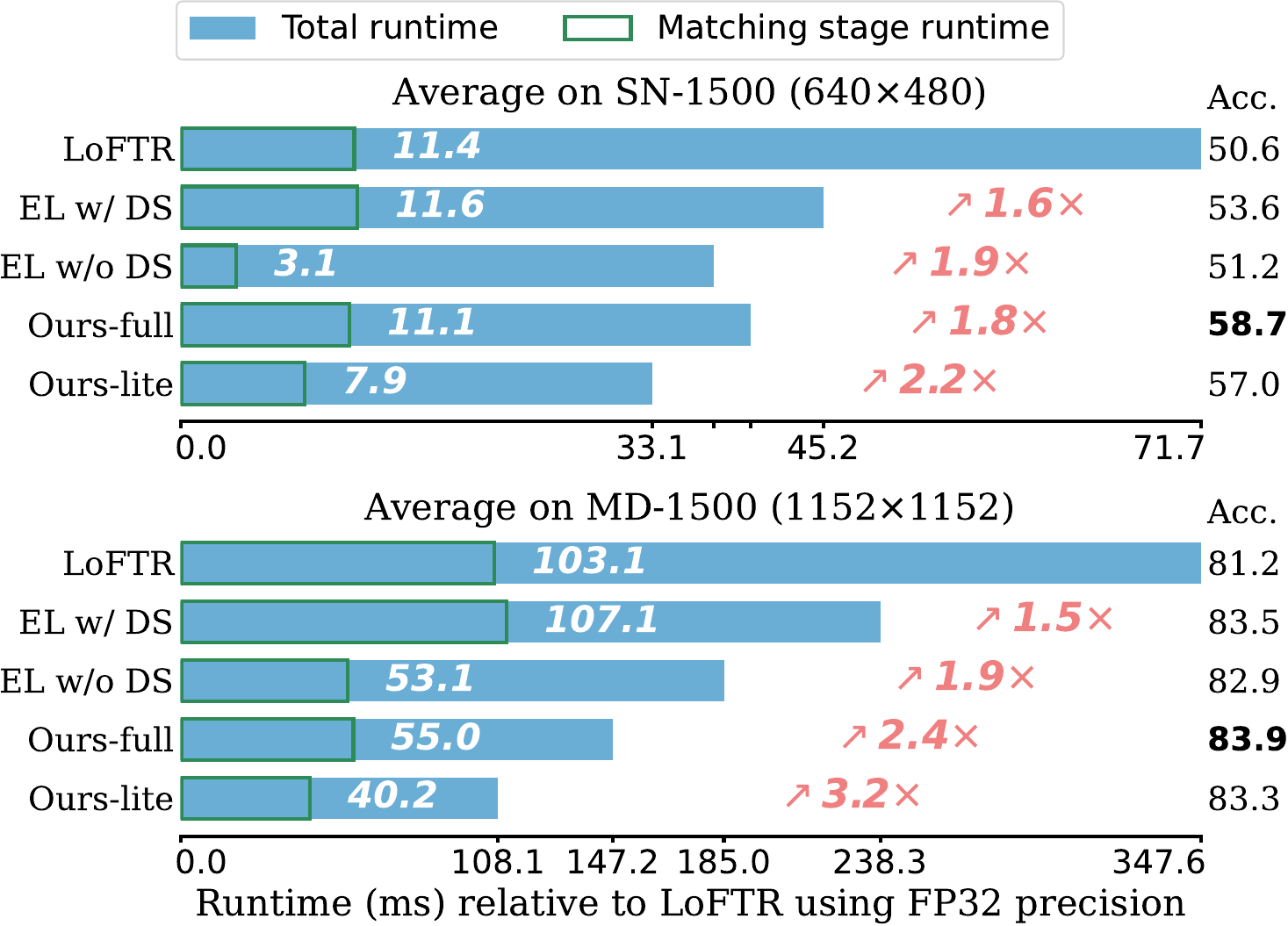}
\caption{
\textbf{Matching Accuracy and Efficiency Comparisons.}
The runtime and AUC@20$\degree$ accuracy are measured across two benchmarks and compared with LoFTR and ELoFTR (EL). The acceleration gains of ELoFTR diminish with increasing resolution since the matching stage occupies a significant portion of runtime. The novel cascaded matching pipeline is more efficient and robust than merely removing the dual-softmax (DS) operator.
}
\label{fig:teaser}
\end{figure}

To tackle this issue, LoFTR~\cite{sun2021loftr} proposes a semi-dense feature matching pipeline that treats each token in the coarse feature map as a potential matching candidate, thereby replacing the feature detection stage. LoFTR enhances robustness under such challenging scenarios by leveraging texture and relative position cues. As a trade-off, the dense interactions among numerous tokens lead to substantial computational costs compared to sparse methods.

\begin{figure}[!t]
\centering
\begin{subfigure}{1.0\linewidth}
\centering
\includegraphics[width=0.95\linewidth]{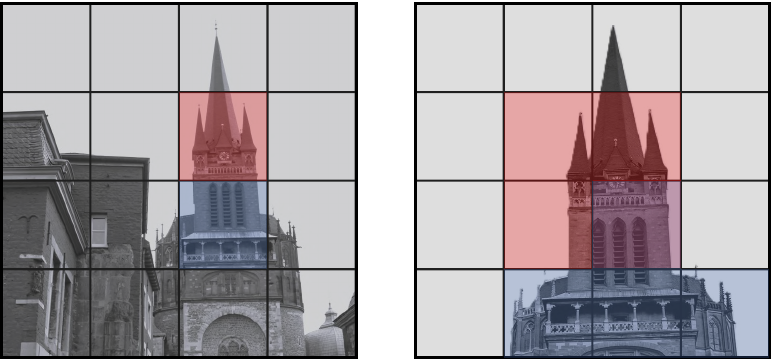}
\caption{one-to-many matching with global search}
\label{fig:schematic_diagram_one_to_many}
\end{subfigure}
\begin{subfigure}{1.0\linewidth}
\centering
\includegraphics[width=0.95\linewidth]{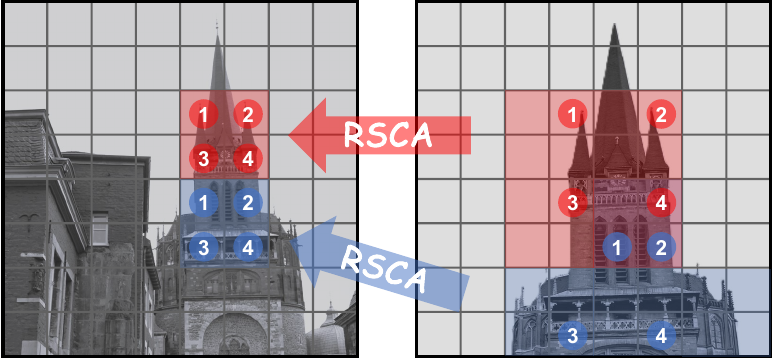}
\caption{one-to-one matching with prior guidance}
\label{fig:schematic_diagram_one_to_one}
\end{subfigure}
\caption{
\textbf{Schematic Diagram of Cascaded Matching.}
(a) One-to-many correspondence priors are selected with global search at a coarse scale and shown as same-colored patches (\textcolor{Purple}{purple} denotes potential common priors). (b) One-to-one matches are determined with prior guidance at the target scale and shown as same-numbered patches, with a region-based attention mechanism, RSCA, applied only at prior token positions.
}
\label{fig:schematic_diagram}
\end{figure}

Follow-up work~\cite{chen2022aspanformer,tang2022quadtree,chen2024affinebased} primarily focused on addressing the limited representational capacity of LoFTR by introducing enhanced feature interaction modules, often at the cost of increased runtime. Recently, ELoFTR~\cite{wang2024efficient} incorporated an aggregated attention mechanism that operates on adaptively selected tokens to improve efficiency. However, we observe that the matching pipeline used by these methods may impose a bottleneck on further advancement. Specifically, the feature maps processed during the matching stage contain excessive tokens, leading to potential latency issues. As shown in Fig.~\ref{fig:teaser}, we evaluate the runtime of our model across two benchmarks at different resolutions and compare it with LoFTR and ELoFTR, with the accuracy of AUC@20$\degree$ reported. As the resolution increases, the matching stage~(\protect\scalerel*{\includegraphics{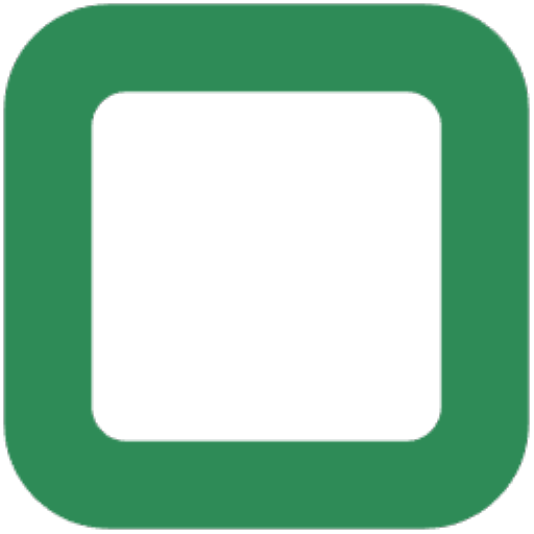}}{B}) of the two methods consistently occupies a substantial portion of the total runtime~(\protect\scalerel*{\includegraphics{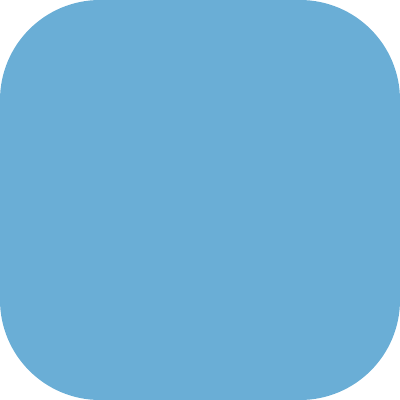}}{B}), which diminishes the expected acceleration gains of ELoFTR. ELoFTR provides a solution by removing the dual-softmax operator, but this leads to a notable accuracy drop across both benchmarks. In contrast, our method achieves significant speedup across all resolutions while also delivering enhanced accuracy, driven by the shift to a more efficient pipeline.

The core guideline for acceleration is to defer primary operations to a coarser scale wherever feasible, thereby reducing the number of tokens processed. To achieve this, we propose a cascaded matching pipeline, CasP, which decomposes the matching stage into two progressive phases. As shown in Fig.~\ref{fig:schematic_diagram}, the pipeline first establishes one-to-many correspondences at a coarser scale as cascaded priors. Then, one-to-one matches at the target scale are determined by leveraging these priors for guidance. The acceleration gains stem from two key factors: \textbf{1)} Instead of conducting a global search across the entire feature map, the second phase focuses only on tokens within the prior areas, eliminating irrelevant computations outside these areas. \textbf{2)} Incorporating high-level features helps reduce the computational costs associated with low-level feature extraction. To ensure more reliable matching confidence, we introduce a region-based selective cross-attention (RSCA) mechanism between the two phases of cascaded matching to enhance feature discriminability among prior candidates. Furthermore, our pipeline adopts a training-inference decoupling strategy, which enhances model representational capacity during training and maximizes inference efficiency.

Building upon the cascaded matching pipeline, we propose a novel semi-dense method integrating advanced modules for enhanced matching accuracy and efficiency. We present two versions of this method, which differ in the number of channels used for low-level feature extraction. Our lite model achieves a speedup of $\sim2.2\times$ and $3.2\times$ at a resolution of 1152 compared to ELoFTR and LoFTR, respectively. Furthermore, an additional boost is attainable by using FP16 precision. In terms of accuracy, our full model achieves state-of-the-art performance in extensive experiments. In particular, the ablation study demonstrates the significant improvement of CasP in cross-domain generalization, which underscores the practical effectiveness of our methods.

Our contributions are summarized as follows:
\begin{itemize}
\item A novel pipeline that leverages cascaded correspondence priors to address the existing efficiency bottleneck.
\item A novel attention mechanism that focuses on prior areas to bridge the two phases of cascaded matching.
\item A novel semi-dense method that integrates advanced modules to deliver superior performance, with strong efficiency and cross-domain generalization for practical applications.
\end{itemize}

\begin{figure*}[!t]
\centering
\includegraphics[width=0.95\linewidth]{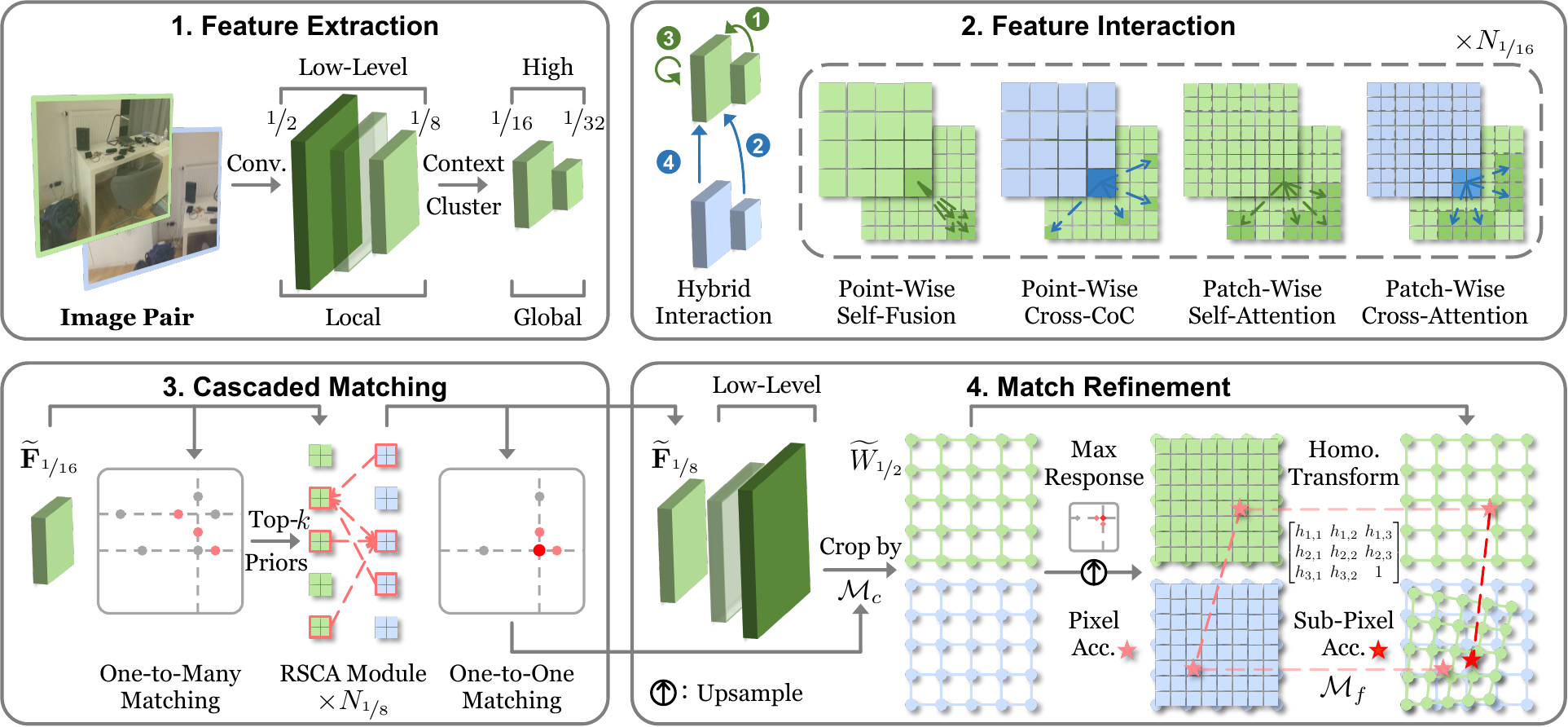}
\caption{
\textbf{Pipeline Overview.}
(1) Low-level ($\nicefrac{1}{2}$ to $\nicefrac{1}{8}$) and high-level ($\nicefrac{1}{16}$ to $\nicefrac{1}{32}$) feature maps are extracted for local and global descriptions, respectively. 
(2) High-level feature maps are transformed by a hybrid interaction module composed of sub-modules that operate at different scales to enhance complementarity.
(3) A cascaded matching module establishes one-to-one correspondences at the $\nicefrac{1}{8}$ scale, with the search range constrained by one-to-many priors identified at the $\nicefrac{1}{16}$ scale.
(4) A two-step homography-based refinement module is applied progressively to reach pixel-level and subpixel-level accuracy.
}
\label{fig:pipeline}
\end{figure*}

    \section{Related Work}
\label{sec:related_work}

\paragraph{Efficient Matching Strategy.}
Sparse methods~\cite{sarlin2020superglue,chen2021learning,shi2022clustergnn} control computational efficiency by adjusting the number of key points extracted by detectors~\cite{lowe2004distinctive,detone2018superpoint,tyszkiewicz2020disk,zhao2022alike,zhao2023aliked,gleize2023silk}. LightGlue~\cite{lindenberger2023lightglue} proposed a pruning scheme that adapts to the varying complexities of image pairs. As the first semi-dense method, LoFTR~\cite{sun2021loftr} employed linear attention~\cite{katharopoulos2020transformers} to ensure manageable computational costs for dense interactions. TopicFM~\cite{giang2023topicfm} introduced a topic-assisted approach, enabling indirect interactions between tokens and fixed-size latent topics. EcoMatcher~\cite{chen2024ecomatcher}, a recently proposed non-transformer-based method, leveraged context clusters to facilitate point-wise interactions with selected anchors. Notably, ELoFTR~\cite{wang2024efficient} adopted a lightweight convolutional neural network (CNN), RepVGG~\cite{ding2021repvgg}, for feature extraction and incorporated an aggregated attention module to perform transformers on reduced tokens.

\vspace{-0.3cm}
\paragraph{Multi-Level Matching Strategy.}
GLU-Net~\cite{truong2020glunet} combined global and local correlation layers to achieve robust and accurate dense correspondence predictions. Building on this global-local architecture, ASpanFormer~\cite{chen2022aspanformer} and AffineFormer~\cite{chen2024affinebased} introduced a multi-level cross-attention mechanism into the semi-dense pipeline. MatchFormer~\cite{wang2022matchformer} proposed an extract-and-match approach that interleaves self- and cross-attention layers at each feature extraction stage. QuadTree~\cite{tang2022quadtree} employed a quadtree-based attention mechanism to focus on relevant regions. PATS~\cite{ni2023pats} adopted a hierarchical framework that sequentially adds and trains each corresponding network at different resolutions. CasMTR~\cite{cao2023improving} incorporated cascade modules after freezing the existing method's feature encoder and coarse attention modules. However, both QuadTree and CasMTR aimed to obtain finer-grained matches based on established coarse correspondences, albeit at the cost of increased runtime.

    \section{Method}
\label{sec:method}

Given a pair of grayscale images, $I^A$ and $I^B$, the existing semi-dense pipeline directly establishes one-to-one matches at a scale of $\nicefrac{1}{8}$ and refines them to sub-pixel accuracy. By contrast, our proposed cascaded pipeline improves efficiency by performing primary operations at a coarser scale of $\nicefrac{1}{16}$ to establish one-to-many correspondence priors, which then guide the subsequent one-to-one matching stage. An overview of our pipeline is shown in Fig.~\ref{fig:pipeline}.

\subsection{Feature Extraction}
\label{sec:feature_extraction}

\paragraph{Low-Level Local Features.} 
A lightweight CNN initially extracts low-level feature maps at scales ranging from $\nicefrac{1}{2}$ to $\nicefrac{1}{8}$ to capture local cues.
The existing pipeline performs a global search across feature maps at the $\nicefrac{1}{8}$ scale to determine coarse matches, which requires a sufficient number of channels to ensure global feature discriminability.
However, this strategy imposes significant throughput bottlenecks for high-resolution inputs because of rapidly increasing computational costs.
Leveraging the proposed cascaded pipeline, our models adopt a modified RepVGG~\cite{ding2021repvgg} architecture with a reduced number of parameters, as shown in Tab.~\ref{tab:low_level_feature_extraction}.

\vspace{-0.3cm}
\paragraph{High-Level Global Features.} 
Since we defer primary operations to a coarser scale, additional down-sampled feature maps, $\mathbf{F}_{\nicefrac{1}{16}}^A$, $\mathbf{F}_{\nicefrac{1}{16}}^B$, $\mathbf{F}_{\nicefrac{1}{32}}^A$, and $\mathbf{F}_{\nicefrac{1}{32}}^B$, are required for subsequent interaction and matching stages.
Rather than applying convolutions for local description, we follow EcoMatcher~\cite{chen2024ecomatcher} and employ the context cluster mechanism~\cite{ma2023image}, referred to as self-CoC, to extract high-level features and enhance contextual understanding with a global receptive field.

The self-CoC module utilizes selected anchors $A$ as proxies to enable indirect point-wise interactions among all feature points $P$.
Specifically, it consists of three main stages: \textit{Clustering}, \textit{Aggregating}, and \textit{Dispatching}. In the \textit{Clustering} stage, each point is allocated to the most similar anchor, forming corresponding clusters $C$. The \textit{Aggregating} stage then updates anchors by aggregating the points within the same clusters. Finally, the \textit{Dispatching} stage propagates contextual information back from anchors to points, completing a round of message exchange. Formally, both points $P$ and anchors $A$ are linearly projected into the similarity and value spaces. The operations for each stage are given by:
\begin{gather}
S=\operatorname{sim}(P^s,A^s),\quad i\in C[j]
\Leftrightarrow j=\operatorname{arg\,max}S[i,:], \label{eq:coc_clustering} \\
\widehat{A}^v[j]=\frac{A^v[j]+\sum_{k\in C[j]}S[k,j]\cdot P^v[k]}{1+\sum_{k\in C[j]}S[k,j]}, \label{eq:coc_aggregating} \\
\widehat{P}^v[i]=\operatorname{sigmoid}(S[i,j])\cdot\widehat{A}^v[j], \label{eq:coc_dispatching}
\end{gather}
where superscripts $s$ and $v$ denote the respective spaces, and $\operatorname{sim}(\cdot,\cdot)$ measures pair-wise similarity. The computational cost remains manageable by controlling the number of $A$.

\begin{table}[!t]
\centering
\resizebox{1.0\columnwidth}{!}{
\setlength\tabcolsep{3pt}
\begin{tabular}{lcccc}
\toprule
\multicolumn{1}{c}{Method} & Type & \#Channels & \#Blocks & \#Params (M) \\
\midrule
LoFTR~\cite{sun2021loftr} & ResNet & [128,196,256] & [2,2,2] & 5.9 \\
ELoFTR~\cite{wang2024efficient} & RepVGG & [64,128,256] & [2,4,14] & 9.5 \\
Ours-full & RepVGG & [64,128,192] & [2,4,4] & 2.0 \\
Ours-lite & RepVGG & [64,64,128] & [2,4,4] & 0.8 \\
\bottomrule
\end{tabular}
}
\caption{
\textbf{Comparison of Low-Level Feature Extraction.} Our models adopt an efficient design benefiting from the novel pipeline.
}
\label{tab:low_level_feature_extraction}
\end{table}

\subsection{Feature Interaction}
\label{sec:feature_interaction}

Following multi-level feature extraction, the interaction stage incorporates cross-view cues to strengthen the similarity of each corresponding token pair. We introduce a hybrid module that comprises two complementary mechanisms.

\vspace{-0.3cm}
\paragraph{Attention Mechanism.}
As the core mechanism of transformers, attention models the point-wise relationships among all involved tokens by measuring the similarities between queries $Q$ and keys $K$ and then obtaining a weighted average of values $V$, which can be generally formulated as:
\begin{equation}
\operatorname{Attn}(Q,K,V)=\operatorname{Softmax}(\sigma_q(Q)\sigma_{k}(K)^T)\sigma_{v}(V).
\end{equation}
Although the interaction stage is deferred to the feature maps at the $\nicefrac{1}{16}$ scale, vanilla attention, where $\sigma_{q}$, $\sigma_{k}$ and $\sigma_{v}$ are identity mappings, still incurs high computational costs. Inspired by ELoFTR~\cite{wang2024efficient}, aggregated attention is employed to down-sample tokens into patches by setting $\sigma_{q}$ as a depthwise convolution layer and $\sigma_{k}$ and $\sigma_{v}$ as max-pooling layers, with kernel size and stride both set to 2. Consequently, the actual interaction is conducted at the $\nicefrac{1}{32}$ scale.

\vspace{-0.3cm}
\paragraph{Cross-CoC Mechanism.} 
The down-sampling of the original tokens in aggregated attention serves as an effective strategy for reducing complexity. However, it sacrifices point-to-point modeling in favor of a patch-to-patch approach. This trade-off may compromise the ability to capture token-level details and potentially impact performance. To address this limitation, we adopt the cross-CoC mechanism from EcoMatcher~\cite{chen2024ecomatcher}, which utilizes coarser-grained tokens from $\mathbf{F}_{\nicefrac{1}{32}}^A$ and $\mathbf{F}_{\nicefrac{1}{32}}^B$ as the selected anchors $\widehat{A}^v$ in Eq.~(\ref{eq:coc_dispatching}). This process facilitates indirect point-wise interactions at the $\nicefrac{1}{16}$ scale, thereby complementing aggregated attention at that processing scale. Moreover, a fusion module is incorporated to enable the exchange of local information across feature maps at the $\nicefrac{1}{16}$ and $\nicefrac{1}{32}$ scales.

\vspace{-0.3cm}
\paragraph{Hybrid Interaction Module.}
The hybrid interaction module is constructed according to the order specified in Fig.~\ref{fig:pipeline} and is repeated $N_{\nicefrac{1}{16}}$ times to generate the transformed feature maps, $\mathbf{\widetilde{F}}_{\nicefrac{1}{16}}^A$ and $\mathbf{\widetilde{F}}_{\nicefrac{1}{16}}^B$. This module not only maximizes computational efficiency but also enhances representational capacity by enabling interactions across different scales.

\begin{figure*}[!t]
\centering
\includegraphics[width=0.95\linewidth]{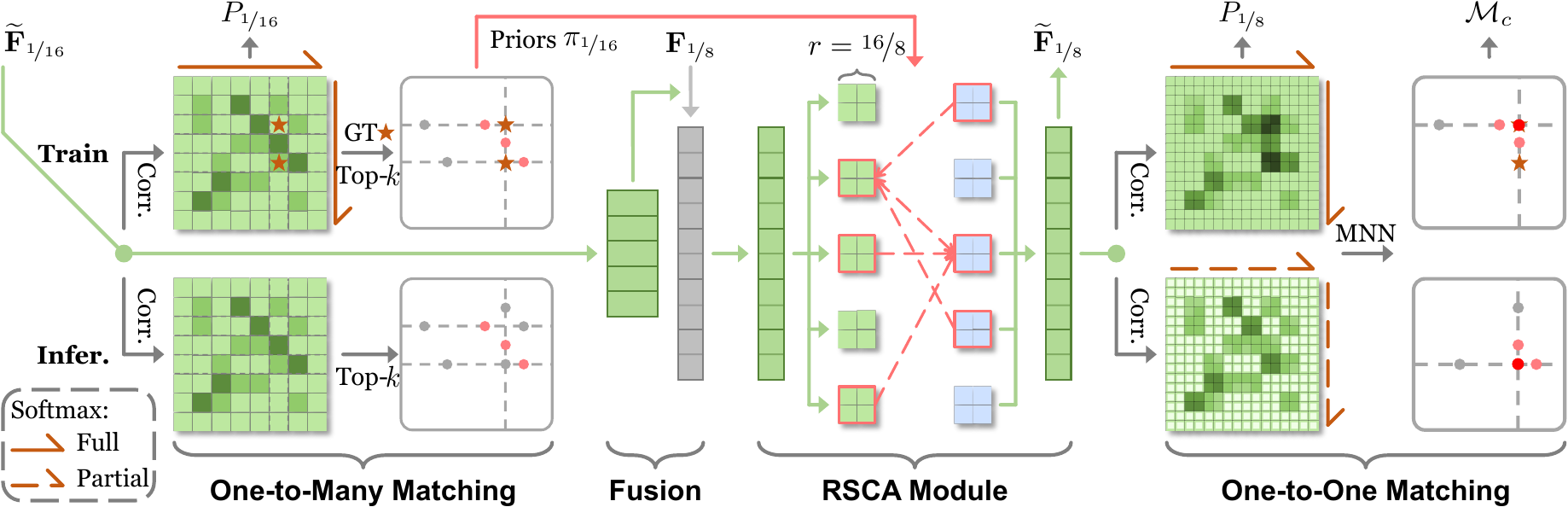}
\caption{
\textbf{Training-Inference Decoupling Cascaded Matching.}
During \textbf{training}, the top-$k$ priors selected by one-to-many matching include ground-truth correspondences for RSCA learning, and the DS operator is applied in both matching stages for supervision. During \textbf{inference}, one-to-many matching omits this step, whereas one-to-one matching applies partial softmax instead of the full version.
}
\label{fig:cascaded_matching}
\end{figure*}

\subsection{Cascaded Matching}
\label{sec:cascaded_matching}

Existing semi-dense methods~\cite{wang2022matchformer,tang2022quadtree,chen2022aspanformer,chen2024ecomatcher,chen2024affinebased} typically follow the LoFTR~\cite{sun2021loftr} pipeline and apply a dual-softmax (DS) operator across both dimensions of the score matrix to filter out low-confidence matches. As evidenced by Fig.~\ref{fig:teaser}, this stage significantly increases the runtime, particularly for high-resolution inputs. ELoFTR~\cite{wang2024efficient} offers a simple solution by directly using the raw score matrix. While this strategy improves efficiency, it compromises robustness and generalizability. AdaMatcher~\cite{huang2023adaptive} introduces a many-to-one assignment mechanism to address scale inconsistencies, but not for efficiency. To pursue a comprehensive solution, we propose a cascaded matching module, as illustrated in Fig.~\ref{fig:cascaded_matching}.

\vspace{-0.3cm}
\paragraph{One-to-Many Matching.}
We first construct the score matrix $S_{\nicefrac{1}{16}}$ from the correlations between $\mathbf{\widetilde{F}}_{\nicefrac{1}{16}}^A$ and $\mathbf{\widetilde{F}}_{\nicefrac{1}{16}}^B$. Our objective is to derive the top-$k$ correspondence priors $\pi_{\nicefrac{1}{16}}$ for each token in both views, which are defined as:
\begin{equation}
\pi_{\nicefrac{1}{16}}^A=\operatorname{arg\,max}_k(S_{\nicefrac{1}{16}}),\quad\pi_{\nicefrac{1}{16}}^B=\operatorname{arg\,max}_k(S_{\nicefrac{1}{16}}^T).
\end{equation}
Assuming one-to-one matches at the $\nicefrac{1}{8}$ scale, we set $k\ge4$ because each token at the $\nicefrac{1}{16}$ scale may correspond to at most $(\nicefrac{16}{8})^2$ tokens in the cross-view feature map.
During \textbf{training}, we apply a DS operator as a differentiable matching layer to $S_{\nicefrac{1}{16}}$, yielding distinctive feature representations and the confidence matrix $P_{\nicefrac{1}{16}}$ for supervision. In addition, we inject one-to-many ground-truth correspondences into $P_{\nicefrac{1}{16}}$ to accelerate convergence in the subsequent one-to-one matching. During \textbf{inference}, the DS operator is omitted because $S_{\nicefrac{1}{16}}$ alone suffices for the top-$k$ selection.

\vspace{-0.3cm}
\paragraph{Region-Based Selective Cross-Attention Mechanism.} 
Prior to the one-to-one matching stage, the previously extracted feature map at the $\nicefrac{1}{8}$ scale is fused with the transformed feature map $\mathbf{\widetilde{F}}_{\nicefrac{1}{16}}$ to inherit cross-view cues, thereby producing $\mathbf{\widetilde{F}}_{\nicefrac{1}{8}}$. Subsequently, we introduce a region-based attention mechanism, RSCA, which allows each token to attend selectively to its correspondence priors in $\pi_{\nicefrac{1}{16}}$.

Specifically, we illustrate this by computing attention from $\mathbf{\widetilde{F}}^B\in\mathbb{R}^{{H^B}\times {W^B}\times c}$ to $\mathbf{\widetilde{F}}^A\in\mathbb{R}^{{H^A}\times {W^A}\times c}$, together with the correspondence priors $\pi^A\in\mathbb{N}^{(H^AW^A/r^2)\times k}$ ($r=\nicefrac{16}{8}$ in our case). Here we omit the subscript for simplicity, and the messages from RSCA are computed as follows:
\begin{align}
&\mathbf{\widehat{F}}^l=\operatorname{Split}_{r}(\mathbf{\widetilde{F}}^l)\in\mathbb{R}^{(H^lW^l/r^2)\times r^2\times c},\quad l\in\{A,B\}, \\
&\mathbf{\widehat{F}}^{A\leftarrow B}=\mathbf{\widehat{F}}^B[\pi^A]\in\mathbb{R}^{(H^AW^A/r^2)\times kr^2\times c}, \\
&\mathbf{\widehat{m}}^{A\leftarrow B}=\operatorname{Attn}(\mathbf{\widehat{F}}^A, \mathbf{\widehat{F}}^{A\leftarrow B}, \mathbf{\widehat{F}}^{A\leftarrow B}), \label{eq:rsca} \\
&\mathbf{m}^{A\leftarrow B}=\operatorname{Merge}_{r}(\mathbf{\widehat{m}}^{A\leftarrow B})\in\mathbb{R}^{H^A\times W^A\times c},
\end{align}
where $\operatorname{Split}_{r}(\cdot)$ partitions the input into cells of size $r\times r$, and $\operatorname{Merge}_{r}(\cdot)$ performs the reverse operation. In Eq.~(\ref{eq:rsca}), the length of each query is $r^2$, and that of each key/value is $k\cdot r^2$. $\mathbf{\widetilde{F}}^A$ is then updated using a feed-forward network that incorporates a convolution for locality modeling, which compensates for the absence of self-attention mechanism:
\begin{gather}
\mathbf{\widetilde{F}}^A=\mathbf{\widetilde{F}}^A+\operatorname{FFN}(\mathbf{\widetilde{F}}^A,\mathbf{m}^{A\leftarrow B}), \\
\operatorname{FFN}(\mathbf{\widetilde{F}},\mathbf{m})=\operatorname{Conv}(\operatorname{GeLU}(\operatorname{Linear}([\mathbf{\widetilde{F}}\|\mathbf{m}]))),
\end{gather}
where $[\cdot\|\cdot]$ represents the concatenation operation. We repeat the RSCA module $N_{\nicefrac{1}{8}}$ times, as shown in Fig.~\ref{fig:pipeline}, to enhance feature discriminability among tokens at prior positions.

\begin{table*}[!t]
\centering
\resizebox{0.9\textwidth}{!}{
\setlength\tabcolsep{12pt}
\begin{tabular}{clcccccccc}
\toprule
\multicolumn{1}{c}{\multirow{2}{*}{Category}} & \multicolumn{1}{c}{\multirow{2}{*}{Method}} & \multicolumn{3}{c}{AUC on MD-1500 $\uparrow$} & \multicolumn{3}{c}{AUC on SN-1500 $\uparrow$} & \multicolumn{2}{c}{Average Runtime~(ms) $\downarrow$} \\
\cmidrule(lr){3-5} 
\cmidrule(lr){6-8} 
\cmidrule(lr){9-10} 
\multicolumn{2}{c}{} & @5$\degree$ & @10$\degree$ & @20$\degree$ & @5$\degree$ & @10$\degree$ & @20$\degree$ & on MD-1500 & on SN-1500 \\
\midrule
\multirow{2}{*}{Sparse} & \uwave{SP}~\cite{detone2018superpoint} + \underline{SG}~\cite{sarlin2020superglue} & 49.7 & 67.1 & 80.6 & 17.4 & 33.9 & 49.5 & \uwave{51.9} + \underline{72.0} & \uwave{36.7} + \underline{72.0} \\
\multicolumn{1}{c}{} & \uwave{SP}~\cite{detone2018superpoint} + \underline{LG}~\cite{lindenberger2023lightglue} & 49.9 & 67.0 & 80.1 & 17.7 & 34.6 & 51.2 & \uwave{51.9} + \underline{44.0} & \uwave{36.7} + \underline{44.0} \\
\midrule
\multirow{7}{*}{Semi-Dense} & \multicolumn{1}{l}{LoFTR~\cite{sun2021loftr}} & 52.8 & 69.2 & 81.2 & 16.9 & 33.6 & 50.6 & 347.6 & 71.7 \\
\multicolumn{1}{c}{} & \multicolumn{1}{l}{QuadTree~\cite{tang2022quadtree}} & 54.6 & 70.5 & 82.2 & 19.0 & 37.3 & 53.5 & 506.4 & 128.5 \\
\multicolumn{1}{c}{} & \multicolumn{1}{l}{ASpanFormer~\cite{chen2022aspanformer}} & 55.3 & 71.5 & 83.1 & 19.6 & 37.7 & 54.4 & 414.0 & 92.6 \\
\multicolumn{1}{c}{} & \multicolumn{1}{l}{ELoFTR~\cite{wang2024efficient}} & 56.4 & 72.2 & 83.5 & 19.2 & 37.0 & 53.6 & 238.3 / 158.8 & 45.2 / 36.5 \\
\multicolumn{1}{c}{} & \multicolumn{1}{l}{AffineFormer$\dagger$~\cite{chen2024affinebased}} & \first{57.3} & \first{72.8} & \first{84.0} & \second{22.0} & \second{40.9} & \second{58.0} & $\ge$347.6 & $\ge$71.7 \\
\multicolumn{1}{c}{} & \multicolumn{1}{l}{Ours-full} & \second{57.1} & \second{72.7} & \second{83.9} & \first{23.0} & \first{41.6} & \first{58.7} & \second{147.2 / 83.8} & \second{40.1 / 32.3} \\
\multicolumn{1}{c}{} & \multicolumn{1}{l}{Ours-lite} & 55.6 & 71.7 & 83.3 & 21.6 & 40.1 & 57.0 & \first{108.1 / 67.7} & \first{33.1 / 30.7} \\
\midrule
\multirow{2}{*}{Dense} & \multicolumn{1}{l}{DKM~\cite{edstedt2023dkm}} & 60.4 & 74.9 & 85.1 & 26.6 & 47.1 & 64.2 & 1355.6 & 414.8 \\
\multicolumn{1}{c}{} & \multicolumn{1}{l}{ROMA~\cite{edstedt2024roma}} & 62.6 & 76.7 & 86.3 & 28.9 & 50.4 & 68.3 & 1482.5 & 493.2 \\
\bottomrule
\end{tabular}
}
\caption{
\textbf{Relative Pose Estimation Results on MD-1500 and SN-1500 with Standard RANSAC.}
All methods are evaluated using a model trained on outdoor scenes. The AUCs of errors up to 5${\degree}$, 10${\degree}$, and 20${\degree}$, and the average runtime, are reported. For ELoFTR, we compare the runtime using FP32/FP16 precision with its full model. $\dagger$ denotes that the runtime is inferred from the paper since the code is unavailable.
}
\label{tab:relative_pose_estimation_ransac}
\end{table*}

\vspace{-0.3cm}
\paragraph{One-to-One Matching.} 
Similarly, the score matrix $S_{\nicefrac{1}{8}}$ is obtained from the correlations, and a DS operator is applied to provide supervision during \textbf{training}. During \textbf{inference}, the softmax operator is applied only to the key/value tokens that are attended to in the RSCA module for each query token, while the remaining ones are omitted and set to zero. This process, referred to as partial softmax, significantly reduces computational costs. The resulting confidence matrix $P_{\nicefrac{1}{8}}$ is defined as follows:
\begin{equation}
\operatorname{PartialSoftmax}(\mathbf{x}, \pi)[i]=
\dfrac{\chi_\pi(i)\operatorname{exp}(\mathbf{x}[i])}{\sum_{k\in \pi}{\operatorname{exp}(\mathbf{x}[k])}}, \label{eq:partial_softmax}
\end{equation}
\vspace{-0.7cm}
\begin{align}
P_{\nicefrac{1}{8}}[i,j]=&\operatorname{PartialSoftmax}(S_{\nicefrac{1}{8}}[i], \phi_r(\pi_{\nicefrac{1}{16}}^A)[i])[j]\odot \notag \\
&\operatorname{PartialSoftmax}({S_{\nicefrac{1}{8}}^T}[j], \phi_r(\pi_{\nicefrac{1}{16}}^B)[j])[i],
\end{align}
where $\chi_\pi(\cdot)$ denotes the indicator function and $\phi_r(\cdot)$ maps each position at the $\nicefrac{1}{16}$ scale to $r^2$ corresponding positions at the $\nicefrac{1}{8}$ scale. Coarse matches are filtered based on a predefined threshold $\theta$ over $P_{\nicefrac{1}{8}}$. The mutual-nearest-neighbor (MNN) criterion is then applied for one-to-one matching, forming $\mathcal{M}_c$. Note that Eq.~(\ref{eq:partial_softmax}) implies that each selected match $(i,j)$ must be derived from the correspondence priors on both sides, which can be formulated as follows:
\begin{equation}
j\in\phi_r(\pi_{\nicefrac{1}{16}}^A)[i]\quad\text{and}\quad i\in\phi_r(\pi_{\nicefrac{1}{16}}^B)[j].
\end{equation}

\subsection{Match Refinement}
\label{sec:match_refinement}

For each match in $\mathcal{M}_c$, a point-to-point correspondence is established at the $\nicefrac{1}{8}$ scale, along with a patch-to-patch correspondence at the original resolution. To refine coarse matches, local patches are first extracted, followed by a two-stage homography-based module for sub-pixel accuracy.

\vspace{-0.3cm}
\paragraph{Local Patch Extraction.}
As with obtaining $\mathbf{\widetilde{F}}_{\nicefrac{1}{8}}$, the feature maps at the $\nicefrac{1}{2}$ and $\nicefrac{1}{4}$ scales are progressively fused in an FPN-like manner. Local patches at the $\nicefrac{1}{2}$ scale, denoted as $\mathbf{\widetilde{W}}_{\nicefrac{1}{2}}$, are then cropped using a $w\times w$ window centered on each coarse match for subsequent refinement.

\vspace{-0.3cm}
\paragraph{Two-Stage Homography.}
In the first stage, we upsample $\mathbf{\widetilde{W}}_{\nicefrac{1}{2}}$ to the original resolution to establish pixel-level correspondences by selecting the maximum response from the correlations. Rather than employing the regression-by-expectation strategy~\cite{sun2021loftr,wang2024efficient} to achieve sub-pixel accuracy, we draw inspiration from HomoMatcher~\cite{wang2025homomatcher} and model the transformations between $\mathbf{\widetilde{W}}_{\nicefrac{1}{2}}^A$ and $\mathbf{\widetilde{W}}_{\nicefrac{1}{2}}^B$ as rigid homographies, disregarding deformable regions. Unlike HomoMatcher’s use of a fixed central location, our method leverages pixel-level correspondences as well-estimated initial positions to enhance accuracy.

\subsection{Supervision}
\label{sec:supervision}

\paragraph{Coarse Supervision.} 
We first construct the one-hot 4D ground truth matrix $M_{\nicefrac{1}{8}}^{gt}$ at the $\nicefrac{1}{8}$ scale by establishing one-to-one correspondences between $I^A$ and $I^B$ using camera poses and depth maps. We then extract the supervision set $\mathcal{M}_{\nicefrac{1}{8}}^{gt}$ by selecting non-zero element positions and converting them into index pairs. Next, we downsample $M_{\nicefrac{1}{8}}^{gt}$ via max-pooling across each dimension to obtain $M_{\nicefrac{1}{16}}^{gt}$ and form $\mathcal{M}_{\nicefrac{1}{16}}^{gt}$ in the same manner. Finally, we define the coarse loss as the negative log-likelihood on the confidence matrix $P_{\nicefrac{1}{s}}$, where $s\in{8,16}$, as follows:

\begin{equation}
L_{\nicefrac{1}{s}}^{c}=-\tfrac{1}{\# \mathcal{M}_{\nicefrac{1}{s}}^{gt}}\sum\nolimits_{(i,j)\in\mathcal{M}_{\nicefrac{1}{s}}^{gt}}\operatorname{log}(P_{\nicefrac{1}{s}}[i,j]). \label{eq:neg_log_likelihood_loss}
\end{equation}

\vspace{-0.3cm}
\paragraph{Fine Supervision.} The pixel-level supervision set $\mathcal{M}_{\nicefrac{1}{1}}^{gt}$ and loss $L_{\nicefrac{1}{1}}^{f}$ can be defined similarly as above. The sub-pixel loss $L_{\text{sub}}^f$ is calculated as a $\ell_2$ loss between the warped positions and the ground truth.

\vspace{-0.3cm}
\paragraph{Total Loss.} The total loss $L$ is formulated as a linear combination of each term mentioned above:
\begin{gather}
L=\lambda_1 L_{\nicefrac{1}{16}}^c+\lambda_2 L_{\nicefrac{1}{8}}^c+\lambda_3 L_{\nicefrac{1}{1}}^f+\lambda_4 L_{\text{sub}}^f.
\end{gather}

    \section{Experiments}
\label{sec:experiments}

Unless otherwise stated, all methods are evaluated by default on a single NVIDIA V100 GPU using FP32 precision. The \colorbox{first}{\bf first} and \colorbox{second}{second} results are highlighted.

\subsection{Implementation Details}
As shown in Tab.~\ref{tab:low_level_feature_extraction}, the only difference between our full and lite models is the number of channels for low-level feature extraction. The number of channels for high-level feature extraction and one-to-many matching is set to 256. $k=8$ correspondence priors are selected for one-to-one matching. The hybrid interaction module and the RSCA module are repeated $N_{\nicefrac{1}{16}}=2$ and $N_{\nicefrac{1}{8}}=2$ times, respectively. The window size $w$ for local patch extraction is set to 5. The loss weights are set to $\lambda_1=0.5$, $\lambda_2=0.5$, $\lambda_3=0.25$, and $\lambda_4=1.0$. Both models are trained on MegaDepth~\cite{li2018megadepth} using 8 NVIDIA V100 GPUs with a batch size of 8 for 30 epochs.

\begin{figure*}[!t]
\centering
\includegraphics[width=0.95\textwidth]{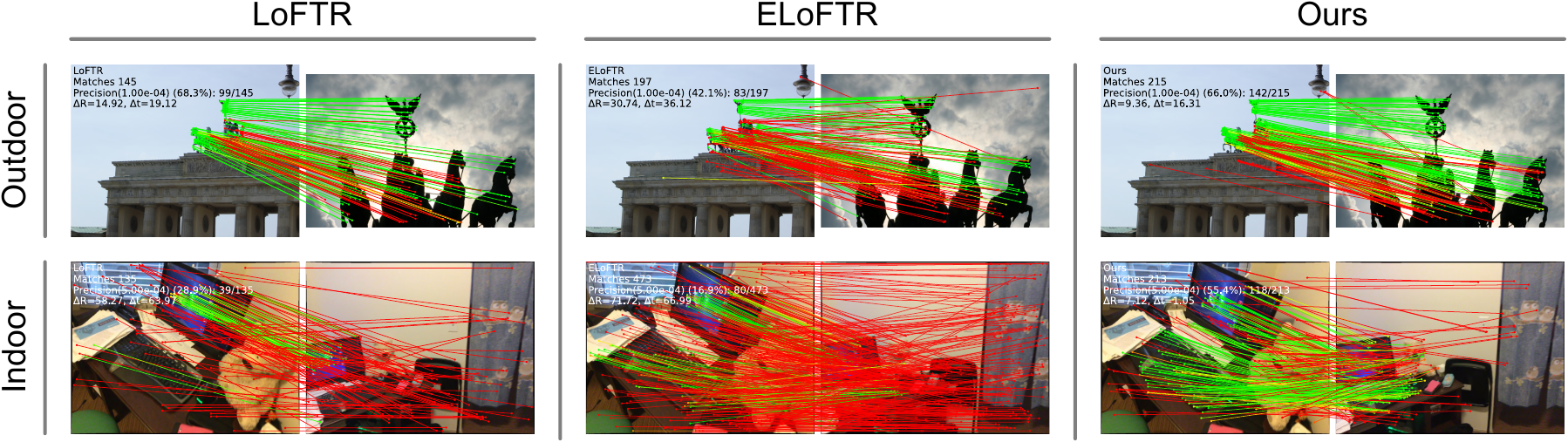}
\caption{
\textbf{Qualitative Results.} Two challenging image pairs are selected for qualitative analysis and compared with LoFTR and ELoFTR. One pair lacks texture details due to illumination changes, while the other undergoes significant viewpoint variations.
}
\label{fig:qualitative_result}
\end{figure*}

\subsection{Relative Pose Estimation}
\label{sec:relative_pose_estimation}

\paragraph{Datasets.}
MD-1500 and SN-1500 are widely adopted benchmarks for estimating relative pose in outdoor and indoor scenes, respectively. The MegaDepth~\cite{li2018megadepth} dataset contains around 130K images that correspond to 196 sparse 3D models reconstructed using COLMAP~\cite{schonberger2016structurefrommotion}. MD-1500, selected by LoFTR~\cite{sun2021loftr}, contains image pairs exhibiting changes in viewpoint and illumination from the ``St. Peter’s Square'' and ``Brandenburger Tor''. The ScanNet~\cite{dai2017scannet} dataset is richly annotated with 3D camera poses and contains 2.5M views from 1,613 indoor scans. SN-1500, selected by SuperGlue~\cite{sarlin2020superglue}, consists of image pairs from scenes with low texture and repetitive patterns.

\vspace{-0.3cm}
\paragraph{Evaluation Protocol.}
We evaluate all methods on both benchmarks using a model trained on outdoor scenes, assessing in-domain and cross-domain performance. Matching accuracy is reported as the area under the curve (AUC) of the relative pose error at various thresholds, and efficiency is measured by the average runtime across benchmarks to reveal resolution effects. All images are resized to align with the corresponding accuracy metrics. Standard RANSAC is used as a common estimator, with a uniform threshold of 0.5 pixels applied in both accuracy and efficiency evaluations.

\vspace{-0.3cm}
\paragraph{Results.}
As shown in Tab.~\ref{tab:relative_pose_estimation_ransac}, the proposed method demonstrates highly competitive performance on both in-domain and cross-domain benchmarks across all categories. In terms of accuracy, our full model achieves results comparable to AffineFormer~\cite{chen2024affinebased}, the best-performing semi-dense method. However, AffineFormer significantly lags behind our method in runtime. Notably, the marked improvement on SN-1500 highlights the strong cross-domain generalization capability of our method, with further validation provided in the ablation study. Regarding efficiency, our lite model achieves a speedup of $\sim$2.2/3.2$\times$ on MD-1500 and $\sim$1.4/2.2$\times$ on SN-1500 compared to ELoFTR/LoFTR using FP32 precision, with additional speedup under FP16 precision.

\begin{table}[!t]
\centering
\resizebox{0.8\columnwidth}{!}{
\setlength\tabcolsep{12pt}
\begin{tabular}{lccc}
\toprule
\multicolumn{1}{c}{\multirow{2}{*}{Method}} & \multicolumn{3}{c}{AUC on HPatches $\uparrow$} \\
\cmidrule{2-4} 
\multicolumn{1}{c}{} & @3px & @5px & @10px \\
\midrule
SP~\cite{detone2018superpoint} + SG~\cite{sarlin2020superglue} & 53.9 & 68.3 & 81.7 \\
LoFTR~\cite{sun2021loftr} & 65.9 & 75.6 &  84.6 \\
ELoFTR~\cite{wang2024efficient} & 66.5 & 76.4 & 85.5 \\
DKM~\cite{edstedt2023dkm} & \second{71.3} & \first{80.6} & \first{88.5} \\
Ours-full & \first{71.8} & \first{80.6} & \second{88.0} \\
\bottomrule
\end{tabular}
}
\caption{
\textbf{Homography Estimation Results on HPatches.}
}
\label{tab:homography_estimation_hpatches}
\end{table}

\subsection{Homography Estimation}
\label{sec:homography_estimation}

\paragraph{Datasets.}
HPatches~\cite{balntas2017hpatches} is a well-established benchmark for homography estimation. It contains 108 sequences, each consisting of 5 image pairs with viewpoint or illumination changes, along with their corresponding ground truth.

\vspace{-0.3cm}
\paragraph{Evaluation Protocol.}
We follow previous work by reporting the AUCs of the mean reprojection error for the four corner points warped by the estimated homography at different thresholds. Similarly, the standard RANSAC solver with a threshold of 2 pixels is used to estimate the homography.

\vspace{-0.3cm}
\paragraph{Results.}
As shown in Tab.~\ref{tab:homography_estimation_hpatches}, the proposed method demonstrates significant improvements in matching accuracy compared to all baseline methods. Notably, it achieves results comparable to the dense method DKM~\cite{edstedt2023dkm}, highlighting the superiority of the two-stage homography-based refinement module in achieving sub-pixel accuracy.

\begin{table}[!t]
\centering
\resizebox{1.0\columnwidth}{!}{
\setlength\tabcolsep{12pt}
\begin{tabular}{lcc}
\toprule
\multicolumn{1}{c}{\multirow{2}{*}{Method}} & Day & Night \\ 
\cmidrule(lr){2-3}
& \multicolumn{2}{c}{(0.25m,2$\degree$) / (0.5m,5$\degree$) / (1.0m,10$\degree$) $\uparrow$}\\ 
\midrule
LoFTR~\cite{sun2021loftr} & 88.7 / 95.6 / 99.0 & 78.5 / 90.6 / 99.0 \\
TopicFM~\cite{giang2023topicfm} & 90.2 / 95.9 / 98.9 & 77.5 / 91.1 / 99.5 \\
ASpanFormer~\cite{chen2022aspanformer} & 89.4 / 95.6 / 99.0 & 77.5 / 91.6 / 99.5 \\
ELoFTR~\cite{wang2024efficient} & 89.6 / 96.2 / 99.0 & 77.0 / 91.1 / 99.5 \\
Ours-full & 89.2 / 96.1 / 98.9 & 78.0 / 91.6 / 99.5 \\
\bottomrule
\end{tabular}
}
\caption{
\textbf{Visual Localization Results on Aachen Day-Night v1.1.}
}
\label{tab:visual_localization_aachen11}
\end{table}

\begin{table}[!t]
\centering
\resizebox{1.0\columnwidth}{!}{
\setlength\tabcolsep{12pt}
\begin{tabular}{lcc}
\toprule
\multicolumn{1}{c}{\multirow{2}{*}{Method}} & DUC1 & DUC2 \\ 
\cmidrule(lr){2-3}
& \multicolumn{2}{c}{(0.25m,2$\degree$) / (0.5m,5$\degree$) / (1.0m,10$\degree$) $\uparrow$} \\ 
\midrule
LoFTR~\cite{sun2021loftr} & 47.5 / 72.2 / 84.8 & 54.2 / 74.8 / 85.5 \\
TopicFM~\cite{giang2023topicfm} & 52.0 / 74.7 / 87.4 & 53.4 / 74.8 / 83.2 \\
ASpanFormer~\cite{chen2022aspanformer} & 51.5 / 73.7 / 86.0 & 55.0 / 74.0 / 81.7\\
ELoFTR~\cite{wang2024efficient} & 52.0 / 74.7 / 86.9 & 58.0 / 80.9 / 89.3 \\
Ours-full & 52.0 / 77.3 / 86.4 & 55.0 / 80.2 / 84.0 \\
\bottomrule
\end{tabular}
}
\caption{
\textbf{Visual Localization Results on InLoc.}
}
\label{tab:visual_localization_inloc}
\end{table}

\subsection{Visual Localization}
\label{sec:visual_localization}

\paragraph{Datasets.} 
Another major application is estimating 6-DoF camera poses relative to a known 3D scene, commonly referred to as visual localization. The Aachen Day-Night v1.1~\cite{sattler2018benchmarking} is a challenging outdoor dataset that involves significant illumination changes, while InLoc~\cite{taira2018inloc} is an indoor dataset characterized by viewpoint changes and occlusions.

\begin{table*}[!t]
\centering
\resizebox{1.0\textwidth}{!}{
\setlength\tabcolsep{3pt}
\begin{tabular}{lllllllllllll}
\toprule
\multicolumn{1}{c}{\multirow{3}{*}{Method}} & \multicolumn{6}{c}{AUC on ETH3D[O]-3438 $\uparrow$} & \multicolumn{6}{c}{AUC on ETH3D[I]-2131 $\uparrow$} \\
\multicolumn{1}{c}{} & \multicolumn{3}{c}{RANSAC} & \multicolumn{3}{c}{MAGSAC++} & \multicolumn{3}{c}{RANSAC} & \multicolumn{3}{c}{MAGSAC++} \\
\cmidrule(lr){2-4}
\cmidrule(lr){5-7}
\cmidrule(lr){8-10}
\cmidrule(lr){11-13}
\multicolumn{1}{c}{} & \multicolumn{1}{c}{@5$\degree$} & \multicolumn{1}{c}{@10$\degree$} & \multicolumn{1}{c}{@20$\degree$} & \multicolumn{1}{c}{@5$\degree$} & \multicolumn{1}{c}{@10$\degree$} & \multicolumn{1}{c}{@20$\degree$} & \multicolumn{1}{c}{@5$\degree$} & \multicolumn{1}{c}{@10$\degree$} & \multicolumn{1}{c}{@20$\degree$} & \multicolumn{1}{c}{@5$\degree$} & \multicolumn{1}{c}{@10$\degree$} & \multicolumn{1}{c}{@20$\degree$} \\
\midrule
EL w/ DS & 56.7 & 63.2 & 69.1 & 58.2 & 64.5 & 70.2 & 49.1 & 55.0 & 59.4 & 51.3 & 56.8 & 61.1 \\
EL w/o DS & 53.4\rd{$_{-3.3}$} & 60.1\rd{$_{-3.1}$} & 66.3\rd{$_{-2.8}$} & 54.7\rd{$_{-3.5}$} & 61.1\rd{$_{-3.4}$} & 67.3\rd{$_{-2.9}$} & 44.7\rd{$_{-4.4}$} & 50.8\rd{$_{-4.2}$} & 55.7\rd{$_{-3.7}$} & 46.2\rd{$_{-5.1}$} & 52.2\rd{$_{-4.6}$} & 57.0\rd{$_{-4.1}$} \\
 \cdashline{1-13} \noalign{\vskip 0.5ex}
EL+CM-full & 60.1\gr{$_{+3.4}$} & 65.6\gr{$_{+2.4}$} & 70.6\gr{$_{+1.5}$} & 61.8\gr{$_{+3.6}$} & 67.1\gr{$_{+2.6}$} & 71.8\gr{$_{+1.6}$} & 52.3\gr{$_{+3.2}$} & 57.2\gr{$_{+2.2}$} & 60.7\gr{$_{+1.3}$} & 54.3\gr{$_{+3.0}$} & 58.8\gr{$_{+2.0}$} & 62.3\gr{$_{+1.2}$} \\
EL+CM-lite & 58.3\gr{$_{+1.6}$} & 64.1\gr{$_{+0.9}$} & 69.4\gr{$_{+0.3}$} & 60.5\gr{$_{+2.3}$} & 66.0\gr{$_{+1.5}$} & 71.1\gr{$_{+0.9}$} & 50.1\gr{$_{+1.0}$} & 54.9\rd{$_{-0.1}$} & 58.6\rd{$_{-0.8}$} & 52.5\gr{$_{+1.2}$} & 57.0\gr{$_{+0.2}$} & 60.3\rd{$_{-0.8}$} \\
\noalign{\vskip 0.25ex} \cdashline{1-13} \noalign{\vskip 0.5ex}
Ours-full & 61.8\gr{$_{+5.1}$} & 66.8\gr{$_{+3.6}$} & 71.5\gr{$_{+2.4}$} & 63.2\gr{$_{+5.0}$} & 68.0\gr{$_{+3.5}$} & 72.6\gr{$_{+2.4}$} & 56.1\gr{$_{+7.0}$} & 60.7\gr{$_{+5.7}$} & 64.0\gr{$_{+4.6}$} & 58.2\gr{$_{+6.9}$} & 62.6\gr{$_{+5.8}$} & 65.6\gr{$_{+4.5}$} \\
Ours-lite & 60.3\gr{$_{+3.6}$} & 65.9\gr{$_{+2.7}$} & 71.2\gr{$_{+2.1}$} & 62.2\gr{$_{+4.0}$} & 67.5\gr{$_{+3.0}$} & 72.6\gr{$_{+2.4}$} & 54.3\gr{$_{+5.2}$} & 59.2\gr{$_{+4.2}$} & 62.7\gr{$_{+3.3}$} & 56.6\gr{$_{+5.3}$} & 61.2\gr{$_{+4.4}$} & 64.5\gr{$_{+3.4}$} \\
\bottomrule
\end{tabular}
}
\captionof{table}{
{\bf Ablation Study on Cross-Domain Relative Pose Estimation.}
EL refers to ELoFTR, and CM denotes the cascaded matching.
}
\label{tab:accuracy_ablation}
\end{table*}
\begin{table}[!t]
\centering
\resizebox{1.0\columnwidth}{!}{
\setlength\tabcolsep{1pt}
\begin{tabular}{lcccccc}
\toprule
\multicolumn{1}{c}{\multirow{2}{*}{Method}} & \multicolumn{1}{c}{\multirow{2}{*}{\shortstack{\#Params \\ (M) \vspace{-0.25cm}}}} & \multicolumn{1}{c}{\multirow{2}{*}{GMACs $\downarrow$}} & \multicolumn{2}{c}{Runtime (ms) $\downarrow$} & \multicolumn{2}{c}{Mem. (GB) $\downarrow$} \\
\cmidrule(lr){4-5}
\cmidrule(lr){6-7}
\multicolumn{1}{c}{} & \multicolumn{1}{c}{} & \multicolumn{1}{c}{} & \multicolumn{1}{c}{FP32} & \multicolumn{1}{c}{FP16} & \multicolumn{1}{c}{FP32} & \multicolumn{1}{c}{FP16} \\
\midrule
EL w/ DS & 16.0 & 909.1 & 238.3 & 158.8 & 13.4 & 13.6 \\
EL w/o DS & 16.0 & 909.1 & 185.3 & 92.8 & 10.1 & 10.3 \\
\noalign{\vskip 0.25ex} \cdashline{1-7} \noalign{\vskip 0.5ex}
EL+CM-full & 17.6 & 708.1 & 144.6 & 79.4 & 12.6 & 7.4 \\
EL+CM-lite & 14.5 & 382.1 & 107.1 & 64.0 & 11.5 & 6.7 \\
\noalign{\vskip 0.25ex} \cdashline{1-7} \noalign{\vskip 0.5ex}
Ours-full & 16.3 & 691.0 & 147.2 & 83.8 & 9.9 & 5.9 \\
Ours-lite & 13.2 & 365.1 & 108.1 & 67.7 & 9.0 & 5.9 \\
\bottomrule
\end{tabular}
}
\caption{
\bf{Ablation Study on Efficiency.}
}
\label{tab:efficiency_ablation}
\end{table}

\vspace{-0.3cm}
\paragraph{Evaluation Protocol.}
Following prior work, we adopt the feature-based framework HLoc~\cite{sarlin2019coarse} to evaluate the accuracy of multi-view matching in visual localization. We report the percentage of query images with localization errors below the specified angular and distance thresholds.

\vspace{-0.3cm}
\paragraph{Results.}
As shown in Tab.~\ref{tab:visual_localization_aachen11} and ~\ref{tab:visual_localization_inloc}, the proposed method achieves competitive results compared to methods that prioritize accuracy. Considered the most efficient semi-dense method, our method can accelerate the matching stage of this framework by $\sim$2 to 3$\times$ compared to other methods.

\subsection{Understanding CasP}
\label{sec:understanding_casp}

\paragraph{Ablation Study.}
The ablation study primarily addresses two concerns: \textbf{a)} How does ELoFTR~\cite{wang2024efficient} perform when the DS operator is removed to speed up inference? \textbf{b)} As shown in Tab.~\ref{tab:relative_pose_estimation_ransac}, CasP achieves a more significant accuracy gain on SN-1500 than on MD-1500. How can this cross-domain generalization be further validated? To investigate these issues, we select two additional datasets, ETH3D[O] and ETH3D[I], from the zero-shot evaluation benchmark proposed by GIM~\cite{shen2024gim}. These benchmarks represent real-world outdoor and indoor scenes in the ETH3D~\cite{schops2017multiview}, respectively. The longer side of each image is resized to 1152 pixels, and standard RANSAC and MAGSAC++~\cite{barath2020magsac} are employed as estimators. We draw the following conclusions from the results shown in Tab.~\ref{tab:accuracy_ablation} and ~\ref{tab:efficiency_ablation}: \textbf{1)} Removing the DS operator is a trade-off that compromises accuracy, as the matching stage relies solely on descriptor similarity and ignores global confidence. \textbf{2)} To examine the core design of our method, we replace the DS operator in ELoFTR with the cascaded matching module described in Sec.~\ref{sec:cascaded_matching}. Even the lite model performs comparably to or better than the original full model. \textbf{3)} Building upon the novel matching pipeline, integrating additional advanced modules in our method further enhances accuracy. \textbf{4)} Our pipeline requires fewer GMACs, delivers faster runtime, and uses less memory on MD-1500.

\begin{figure}[!t]
\centering
\includegraphics[width=0.95\linewidth]{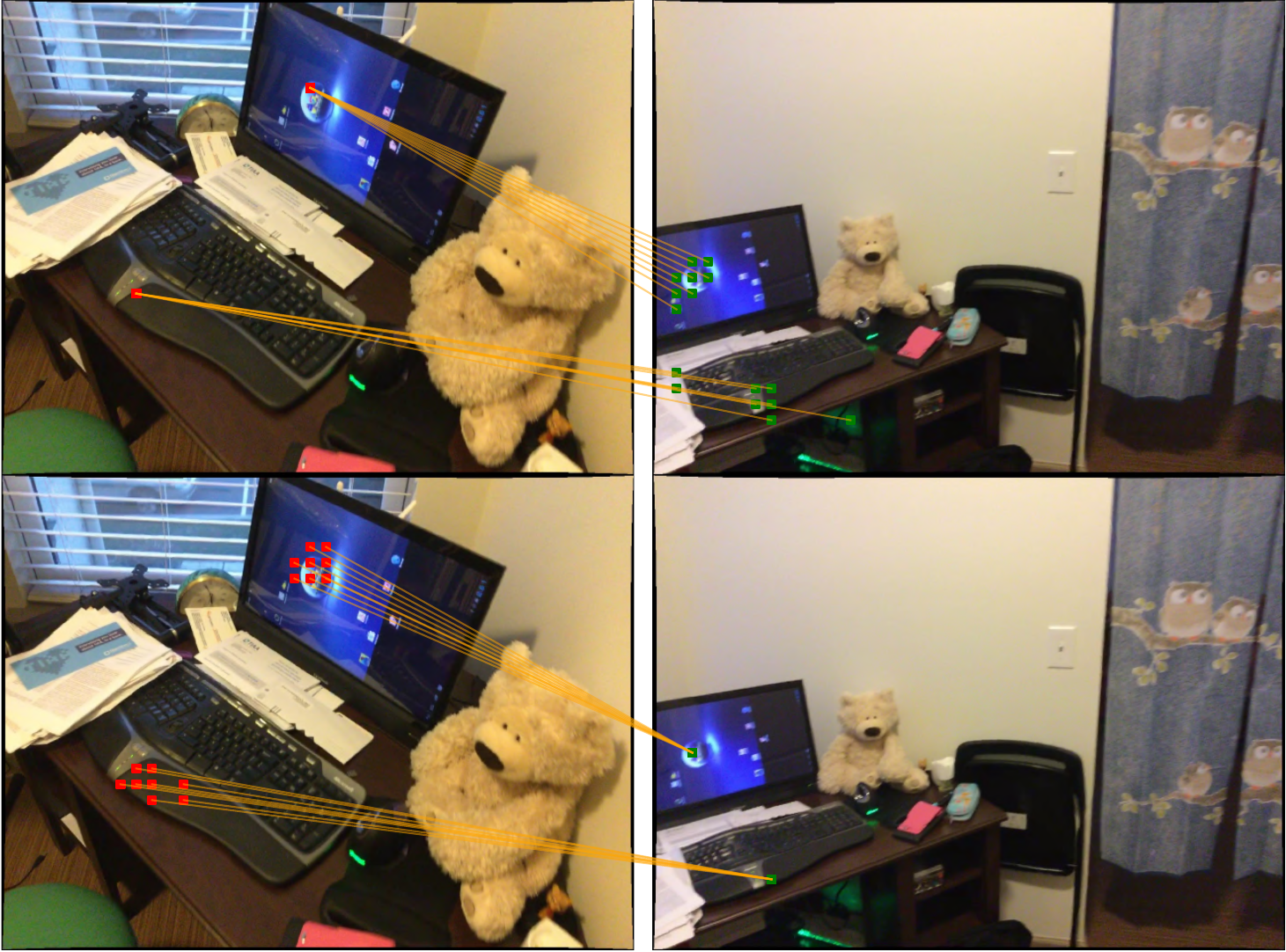}
\caption{
\textbf{Visualization of One-to-Many Matching.}
For each token, $k=8$ priors are displayed in the corresponding image.
}
\label{fig:visualization_of_one_to_matching}
\end{figure}

\vspace{-0.3cm}
\paragraph{Visualization.}
We present qualitative results in Fig.~\ref{fig:qualitative_result} and visualize the one-to-many matching priors in Fig.~\ref{fig:visualization_of_one_to_matching}. These priors significantly facilitate identifying the most probable positions for one-to-one matching in challenging scenarios.

\section{Conclusion}
\label{sec:conclusion}

In this paper, we propose a cascaded matching pipeline to address the efficiency bottleneck of existing methods. Building upon this pipeline, we introduce a novel semi-dense method, CasP, which integrates advanced modules to enhance both matching accuracy and efficiency. Compared to the state-of-the-art method ELoFTR, our method achieves a speedup of $\sim2.2\times$ at a resolution of 1152. Moreover, extensive experiments demonstrate that this novel pipeline significantly contributes to cross-domain generalization. These improvements are crucial for real-world applications, particularly for latency-sensitive and high-robustness tasks.

    \section*{Acknowledgements}
\label{sec:acknowledgements}

This work was supported by the National Key Research and Development Program of China under Grant 2024YFB3909001; the National Natural Science Foundation of China under Grants 42192583, 42030102, and 42471470; the China Railway Group Laboratory Basic Research Project under Grant L2023G014; the Major Special Projects of Guizhou [2022]001; and the Ant Group Research Fund.

    {
        \small
        \bibliographystyle{ieeenat_fullname}
        \bibliography{main}
    }

\end{document}